\documentclass[a4paper, conference]{IEEEtran}
\usepackage[left=1.42cm,right=1.41cm,top=2cm,bottom=4.33cm]{geometry}
\IEEEoverridecommandlockouts

\usepackage{cite}
\usepackage{amsmath,amssymb,amsfonts}
\usepackage{graphicx}
\usepackage{textcomp}
\usepackage[dvipsnames]{xcolor}
\usepackage{algorithm, algpseudocode}

\algnewcommand\algorithmicinput{\textbf{Input:}}
\algnewcommand\INPUT{\item[\algorithmicinput]}
\algnewcommand\algorithmicoutput{\textbf{Output:}}
\algnewcommand\OUTPUT{\item[\algorithmicoutput]}
\algnewcommand{\algorithmicand}{\textbf{ and }}
\algnewcommand{\algorithmicor}{\textbf{ or }}
\algnewcommand{\OR}{\algorithmicor}
\algnewcommand{\AND}{\algorithmicand}
\def\BibTeX{{\rm B\kern-.05em{\sc i\kern-.025em b}\kern-.08em
    T\kern-.1667em\lower.7ex\hbox{E}\kern-.125emX}}
\begin{document}
\title{Multi-Robot Task Allocation for Homogeneous Tasks with Collision Avoidance via Spatial Clustering}

\author{\IEEEauthorblockN{Rathin Chandra Shit}
\IEEEauthorblockA{\textit{IIIT Bhubaneswar} \\
India\\
rathin088@gmail.com}
\and
\IEEEauthorblockN{Sharmila Subudhi}
\IEEEauthorblockA{\textit{Dept. of Computer Science} \\
\textit{Maharaja Sriram Chandra Bhanja Deo University}\\
Baripada, Odisha, India \\
sharmilasubudhi@ieee.org}
}

\maketitle
\begin{abstract}
In this paper, a novel framework is presented that achieves a combined solution based on Multi-Robot Task Allocation (MRTA) and collision avoidance with respect to homogeneous measurement tasks taking place in industrial environments. The spatial clustering we propose offers to simultaneously solve the task allocation problem and deal with collision risks by cutting the workspace into distinguishable operational zones for each robot. To divide task sites and to schedule robot routes within corresponding clusters, we use K-means clustering and the 2-Opt algorithm. The presented framework shows satisfactory performance, where up to 93\% time reduction (1.24s against 17.62s) with a solution quality improvement of up to 7\% compared to the best performing method is demonstrated. Our method also completely eliminates collision points that persist in comparative methods in a most significant sense. Theoretical analysis agrees with the claim that spatial partitioning unifies the apparently disjoint tasks allocation and collision avoidance problems under conditions of many identical tasks to be distributed over sparse geographical areas. Ultimately, the findings in this work are of substantial importance for real world applications where both computational efficiency and operation free from collisions is of paramount importance.

\end{abstract}


\section{Introduction}

The need for Multi Robot Systems (MRS) is increasing in many industries, including agriculture, inspection, search and rescue, and product distribution in warehouses \cite{ju2022review}. Applications such as obtaining inspection data or intervention during disasters in industrial areas present a promising avenue in using coordination among Unmanned Ground Vehicles (UGVs), Unmanned Aerial Vehicles (UAVs), Unmanned Surface Vehicles (USVs) \cite{gielis2022critical}. However, providing security and safety measures in high risk scenarios, like swift gathering of precise and dynamic data during significant incidents, has stood as an open research problem in the last decade \cite{he2024security}. Thus, it is necessary to empower the robots with efficient task allocation algorithms for fine-tuning the decision making process \cite{he2024security, gielis2022critical}. 

The Multi-Robot Task Allocation (MRTA) refers to the problem of arranging multiple mobile robots to perform many tasks under varying constraints to achieve the preset objectives \cite{liang2021secure}. Several market based, optimization based or behavior based strategies exist to resolve MRTA scenarios \cite{gielis2022critical}. A common market-based approach is a bidding process that requires robots to determine an optimal behavior to begin with or some type of behavior to adapt to. Likewise, the behavior based methods allow robots to make independent decisions based on specific behaviors \cite{ju2022review}. For complex problems, heuristics and meta heuristics are common for exact or approximate optimization based approaches \cite{gielis2022critical}.

Collision avoidance is a fundamental challenge in multi-robot operations, especially in the shared workspaces \cite{raibail2022decentralized, park2024survey}. Despite the advantage of shared workspace available for the collaborative robots (cobots), the shared workspace imposes limitations because of possible human operator and robot interference that can create emergency stops and degrade system performance \cite{calzavara2023multi}. 

Therefore, it is of prime concern to put a collision avoidance strategy along with MRTA, since task allocation algorithms might allocate tasks based on close proximity in order to avoid potential conflicts while path or motion planning for multi robot system. 

In this paper, we have developed an MRTA strategy implemented at the task allocation level, where several multiple (homogeneous) tasks are allocated with spatial clustering based collision avoidance technique. The main contributions of this work are:

\begin{itemize}
    \item Integration of Spatial clustering for task allocation and collision avoidance for offering a single solution.
    \item A formal mathematical framework establishing the importance of spatial partitioning as a natural approach for reducing collision risks and simplifying the task allocation problem.
    \item Designing a local routing optimization algorithm for finding out the shortest path among the routes.
    \item Extensive exploration of solution quality tradeoffs and a comprehensive analysis in terms of efficiency gains compared to existing optimization methods.
\end{itemize}

Section \ref{sec2} formalizes the MRTA problem with homogeneous tasks. Section \ref{sec3} details the suggested spatial clustering approach for collision avoidance along with the local routing optimization within clusters. Section \ref{sec4} presents the experiment results and analysis, while Section \ref{sec5} draws the conclusions and future work.

\section{Problem Formulation}\label{sec2}

\subsection{Mathematical Formulation} \label{sec21}
The problem concerns a fleet of mobile sensing robots that must perform a set of measurement tasks at 'sites' in an industrial zone. Thus, we assume that the environment is a known navigation setting (or a grid map). 

Formally, we define:
\begin{itemize}
    \item $A = \{a_1, \dots, a_n\}$: Set of sites, where $a_1$ represents the central depot where robots begin and end their missions.
    \item $R = \{r_1, \dots, r_m\}$: Set of robots.
    \item $M = \{m_1, \dots, m_p\}$: Set of measurement types.
    \item $J = \{j_1, \dots, j_T\}$: Set of tasks, where each task $j_t = (i, q)$ denotes a measurement of type $m_q \in M$ at site $a_i \in A$.
    \item $B_k$: Maximum allowable travel cost (time or energy) for robot $r_k$.
\end{itemize}
Each robot $r_k \in R$ is equipped with sensors corresponding to measurement types $M$. Typically, the travel cost between sites $a_i$ and $a_j$ of robot $r_k$ is computed using Dijkstra's algorithm on the grid map. Thus, aiming to minimize the sum of the costs over all robots forms the objective function (as mentioned in Eq. \eqref{eq1}).
\begin{equation} \label{eq1}
\min \sum_{k=1}^{|R|} \sum_{i=1}^{|A|} \sum_{j=1}^{|A|} c(i,j,k) \cdot x_{ijk}
\end{equation}
where, $x_{ijk}$ is a binary decision variable that equals 1 if robot $r_k$ travels from site $a_i$ to site $a_j$, and 0 otherwise.

\subsection{Task Allocation Constraints} \label{sec22}
The task allocation problem is subject to the following constraints:
\begin{itemize}
    \item Each task must be assigned to exactly one robot.
    \item A robot can perform multiple tasks at a single site simultaneously.
    \item All robots must start and end at the depot.
    \item The total cost for each robot cannot exceed its maximum allowable travel cost $B_k$.
\end{itemize}

In the homogeneous tasks allocation (where all tasks require the same type of measurement), the focus is primarily only on site allocation of tasks. Hence, making the result is more tractable when considering spatial distribution of tasks.

Specifically, it is assumed that each robot is capable of every task and the depot must be clearly distinct from the task locations. However, this separation also plays a significant role in defining the priori areas to partition task locations into few sub regions. Thus, such spatial specifications can be taken into consideration during collision avoidance in task allocation \cite{chen2022graph}.

\subsection{Theoretical Foundation for Spatial Clustering for MRTA}\label{sec23}
Ensuring collision avoidance of multiple agents in a shared space is critical when coordinating multiple agents. The traditional collision avoidance methods are effective, yet generating high computing overhead \cite{chen2022graph}. Therefore, spatial clustering is adopted to detect new errors while continuously monitoring the task allocation.

If we consider multiple homogeneous tasks occurring in a limited geographical area, then we can divide the MRTA problem to clusters. Here, each cluster represents the partitioning of the operational space into distinct regions where a certain robot is assigned to serve the tasks. This can be expressed, in Eq. \eqref{eq2}, as the probability of collision between two robots, $r_i$ and $r_j$.

\begin{equation} \label{eq2}
P(\text{collision}_{i,j}) = \int_{\mathcal{T}} P(r_i \text{ at } \mathbf{x}, t) \cdot P(r_j \text{ at } \mathbf{x}, t) \,d\mathbf{x}\,dt
\end{equation}
where, the spatiotemporal domain of the mission is denoted by $\mathcal{T}$.

The probability distributions can be viewed as almost disjoint when robots are operating in their separate spatial clusters, as mentioned in Eq. \eqref{eq3}. It significantly reduces the overall collision probability and thus can be avoided during the task allocation phase.

\begin{equation} \label{eq3}
P(r_i \text{ at } \mathbf{x}, t) \cdot P(r_j \text{ at } \mathbf{x}, t) \approx 0 \quad \forall \mathbf{x}, t
\end{equation}

To begin with the Spatial Clustering, in this work, initially we have used the K-means clustering for identifying the sites $A$ besides the depots as points that need clustering for effective spatial compact clustering. Then, the system requires the implementation of $|R|$ clusters which directly matches the count of robots. The objective of using K-means coupled with Spatial clustering is to minimize the within-cluster maximum variance (WCM), as mentioned in Eq. \eqref{eq4}.
\begin{equation} \label{eq4}
\min WCM = \sum_{i=1}^{|A|} \| a_i - \mu_{L^{(i)}} \|^2
\end{equation}
where, $\mu_L$ is the mean of the cluster centroids ($L$) of each site $a_i \in A$.

\section{Proposed Approach} \label{sec3}
\subsection{Proposed Collision Avoidance via Spatial Clustering} \label{sec31}
Algorithm \ref{alg1} formalizes the proposed spatial clustering methodology for allocating tasks and preventing collisions in a multi robot system. Each robot receives a spatially organized cluster of tasks from this algorithm to minimize the inter-robot collision risks and simplify the individual route optimization tasks for the robots.

\begin{algorithm}
\caption{Spatial Clustering for MRTA and Collision Avoidance}
\label{alg1}
\begin{algorithmic}[1]
\INPUT{Sites $A = \{a_1, \dots, a_n\}$, Robots $R = \{r_1, \dots, r_m\}$, threshold: $th$}
\OUTPUT{Task assignments and routes for each robot}
\State $A' \gets A \setminus \{a_1\}$ 
\Comment{Exclude depot from clustering}
\State Initialize $m$ centroids $\mu_1, \mu_2, \ldots, \mu_m$ from $A'$
\State $converged \gets false$
\While{not $converged$}
    \State Assign each site to nearest centroid: $$L^{(i)} = \arg\min_j \| a_i - \mu_j \|^2$$
    \State Update centroids: $\mu_j = \frac{1}{|C_j|} \sum_{i: L^{(i)}=j} a_i$
    \If{$\mu_j < th$}
        \State $converged \gets true$
    \EndIf
\EndWhile
\For{each robot $r_k \in R$}
    \State $cluster_k \gets \{a_i \in A' | L^{(i)} = k\}$
    \State $route_k \gets$ OptimizeRoute($\{a_1\} \cup cluster_k \cup \{a_1\}$) 
    \Comment{Start and end at depot}
\EndFor
\State \Return task assignments and routes for each robot
\end{algorithmic}
\end{algorithm}

\subsection{Proposed Local Routing Optimization Algorithm} \label{sec32}

Each robot must determine the best site visiting order within its assigned cluster that both begins and ends at the depot. We have suggested a local search method, known as 2-Opt strategy to determine the shortest path from one place to another. This algorithm enhances an initial tour through an iterative process and replaces two edges by two new edges to form a shorter path, as presented in Algorithm \ref{alg2}. 

\begin{algorithm}
\caption{2-Opt Local Search}
\label{alg2}
\begin{algorithmic}[1]
\INPUT{Initial tour $T$, cost matrix $c$}
\OUTPUT{Optimized tour $T$}
\State $improvement \gets true$
\While{$improvement$}
    \State $improvement \gets false$
    \For{$i \gets 1$ to $|T|-1$}
        \For{$j \gets i+1$ to $|T|$}
            \If{$c_{T[i],T[i+1]} + c_{T[j],T[j+1]} > c_{T[i],T[j]} + c_{T[i+1],T[j+1]}$}
                \State Reverse the segment from position $i+1$ to position $j$ in tour $T$
                \State $improvement \gets true$
            \EndIf
        \EndFor
    \EndFor
\EndWhile
\State \Return $T$
\end{algorithmic}
\end{algorithm}

This algorithm examines pairs of edges $(a_i \to a_{i+1})$ and $(a_j \to a_{j+1})$ from the robot site $A$. It tests all alternative edges like $(a_i \to a_j)$ and $(a_{i+1} \to a_{j+1})$ when looking for sequence improvements. The selection rules guarantee that the new edges form a valid tour with each site must have exactly one incoming and one outgoing connection. The replacement succeeds as long as the new path shows shorter length.

\section{Experimental Evaluation}\label{sec4}
To validate the proposed clustering-based approach for task allocation and collision avoidance for homogeneous tasks, we have conducted a comprehensive simulation study featuring four robots ($R=4$) and fifty distinct sites ($A=50$), including the central depot ($a_1$).

\subsection{Experimental Setup} \label{sec41}
The simulated workspace is of $100 \times 100 m^2$ area where the measurement sites followed a uniform random distribution. Placing the depot at $(0,0)$ ensured its sufficient distance from task locations for preventing clustering interference. The robots maintained a limit of $1$ m/s speed with all the sensory inputs for the measurements. The robots were constrained to travel at $500$ units as the maximum allowed expense.

K-means clustering was with Euclidean distance measure was used to calculate the operational zones for the robots. The implementation of K-means clustering started with $4$ random centroids with $100$ maximum iterations or a reached convergence at $10^{-4}$ tolerance. The 2-Opt algorithm runs separately for each robot within its designated cluster through utilization of clustering outputs and Euclidean distance-based cost calculations.

\subsection{Evaluation Metrics} \label{sec42}
Following metrics are used for evaluating the efficacy of the proposed model.
\begin{itemize}
    \item \textbf{Total Mission Cost} refers to the sum of travel expenses among all robots.
    \item \textbf{Computational Time} denotes the computational solution generation time.
    \item \textbf{Load Balance} signifies the workload distribution consistency among the robots.
    \item \textbf{Collision Risk Assessment} indicates the number of path intersections between different robots' trajectories.
\end{itemize}

\subsection{Results and Analysis} \label{sec43}

\begin{figure}
\centering
\includegraphics[width=0.45\textwidth]{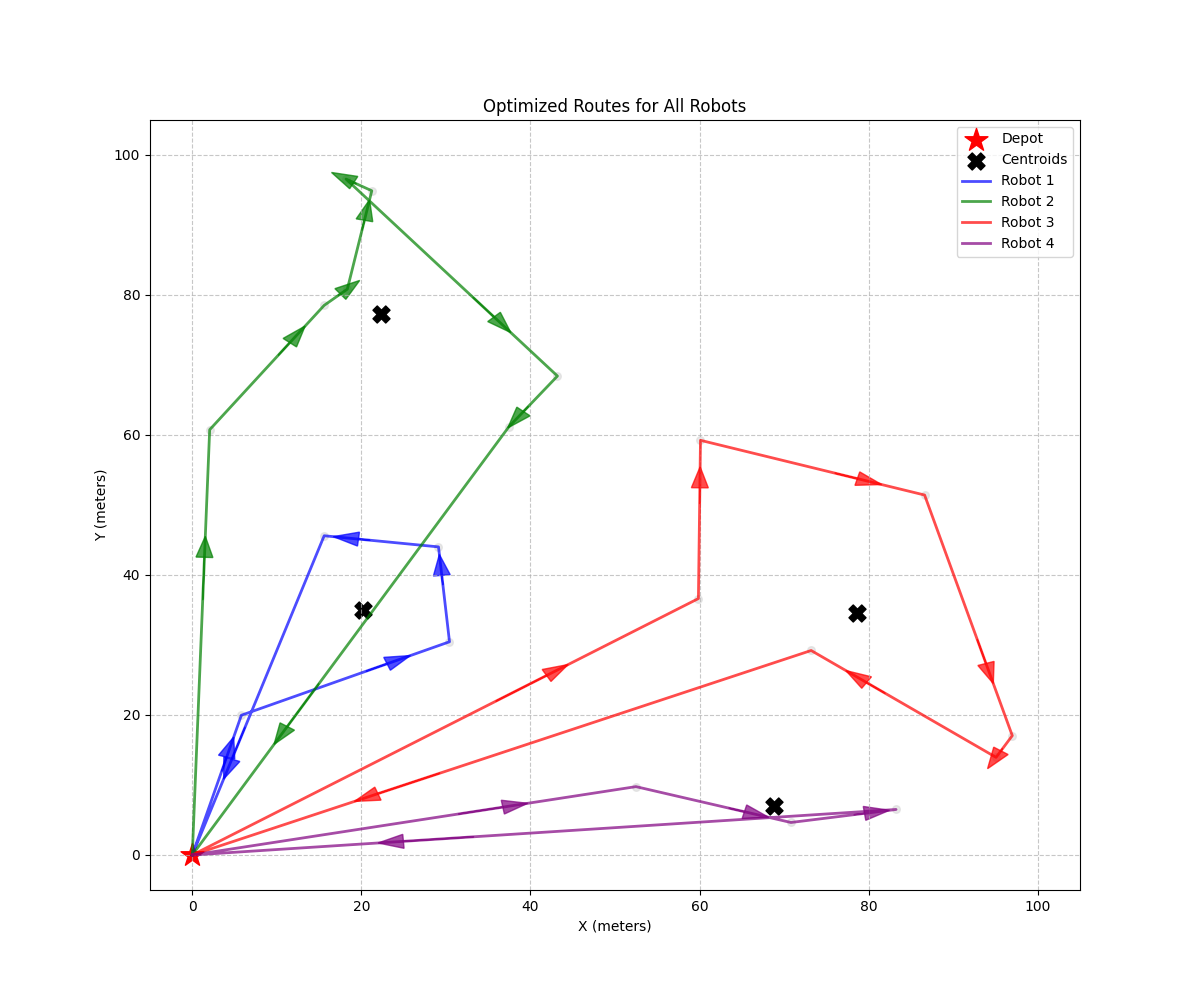}
\caption{Spatial Clustering and Path Visualization of  Robots}
\label{fig:clustering}
\end{figure}

Initially, our clustering based approach partitions the workspace into four distinct operational zones and assigns robot to each cluster. Figure \ref{fig:clustering} presents the mission path of each robot. The dotted lines highlights each robot’s trajectory within its zone. A main observation is that the robotic paths are not intersecting with each other due to the spatial clustering, resulting in collision avoidance. This type of spatial separation naturally minimizes the potential conflicts while ensuring efficient task completion over the whole workspace.

\begin{figure}
\centering
\includegraphics[width=0.45\textwidth]{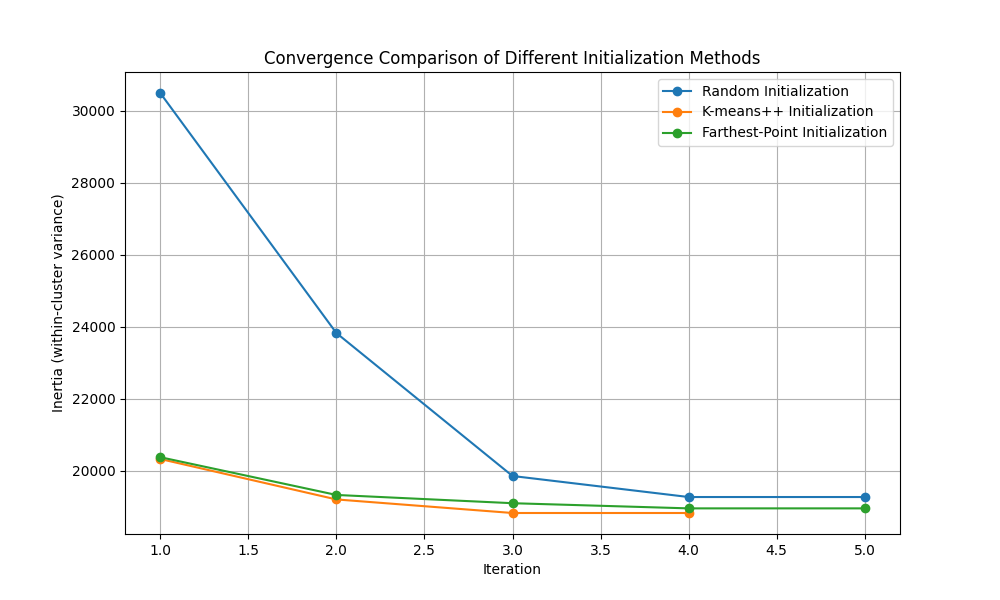}
\caption{Convergence of K-means Algorithm}
\label{fig:kmeans}
\end{figure}

Figure \ref{fig:kmeans} shows the evolution of the K-means algorithm during execution with systematic reduction of the within-cluster maximum variance (WCM). The algorithm reached a stable outcome through eight iterative cycles when processing fifty sites. After clustering, each robot was then optimized route in its assigned cluster using the 2-Opt algorithm. 

\begin{figure}[!htbp]
\centering
\includegraphics[width=0.45\textwidth]{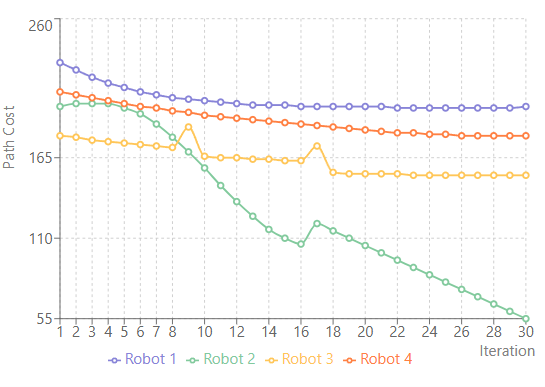}
\caption{Cost Function during 2-Opt Localization for Each Robot Cluster}
\label{fig:2opt}
\end{figure}

Figure \ref{fig:2opt} depicts the evolution of the cost function during 2-Opt localization search. The algorithm overall decreases the cost as it performs the optimization, thus showing that the algorithm is able to escape local minima. 

\subsection{Comparative Performance Analysis} \label{sec44}
We have conducted a comparative performance analysis of our proposed model with three literature \cite{shao2021robot, duan2023research, guo2023efficient}.
\begin{itemize}
    \item Shao used Genetic Algorithm (GA) to determine the optimal path in the multi-robot environment without creating spatial clusters \cite{shao2021robot}.
    \item Duan and Luo developed a GA-based method for path planning of wielding robots \cite{duan2023research}. They measured the motion smoothing level, weighting factors and route distance of the robots to determine the shortest path.
    \item Guo and Yu employed a Greedy method for selecting the unassigned tasks based on the robot's nearest location \cite{guo2023efficient}.
\end{itemize}

Table \ref{tab:comparison} presents a comparative analysis of our approach with three methods \cite{shao2021robot, duan2023research, guo2023efficient}. In computational efficiency, our suggested model has provided significantly better results than direct optimization methods. Duan and Luo's method has a lesser total cost, but took about 14 times longer to complete and eliminated all the collision points. This finding thus substantiates our claim that spatial clustering is an effective collision avoidance solution that can be realized at the task allocation output level. Further, our approach ensured more equitable robot workload distribution through K-means by virtue of its variance-minimizing objective function.

\begin{table}[t]
\centering
\caption{Performance comparison of different task allocation approaches}
\resizebox{\columnwidth}{!}{%
\begin{tabular}{l c c c c}
\hline
\textbf{Method} & \textbf{\begin{tabular}[c]{@{}c@{}}Execution\\Time (s)\end{tabular}} & \textbf{\begin{tabular}[c]{@{}c@{}}Total\\Cost\end{tabular}} & \textbf{\begin{tabular}[c]{@{}c@{}}Load\\Balance\end{tabular}} & \textbf{\begin{tabular}[c]{@{}c@{}}Collision\\Risk\end{tabular}} \\
\hline
Proposed Model & 1.24 & 782.6 & 42.3 & 0 \\
Shao \cite{shao2021robot}& 15.37 & 756.9 & 67.8 & 18 \\
Duan and Luo \cite{duan2023research} & 17.62 & 731.2 & 51.4 & 14 \\
Guo and Yu \cite{guo2023efficient}& 0.08 & 945.3 & 118.7 & 25 \\
\hline
\end{tabular}%
}
\label{tab:comparison}
\end{table}

\subsection{Scalability Analysis} \label{sec45}
\begin{figure}[!htbp]
\centering
\includegraphics[width=0.45\textwidth]{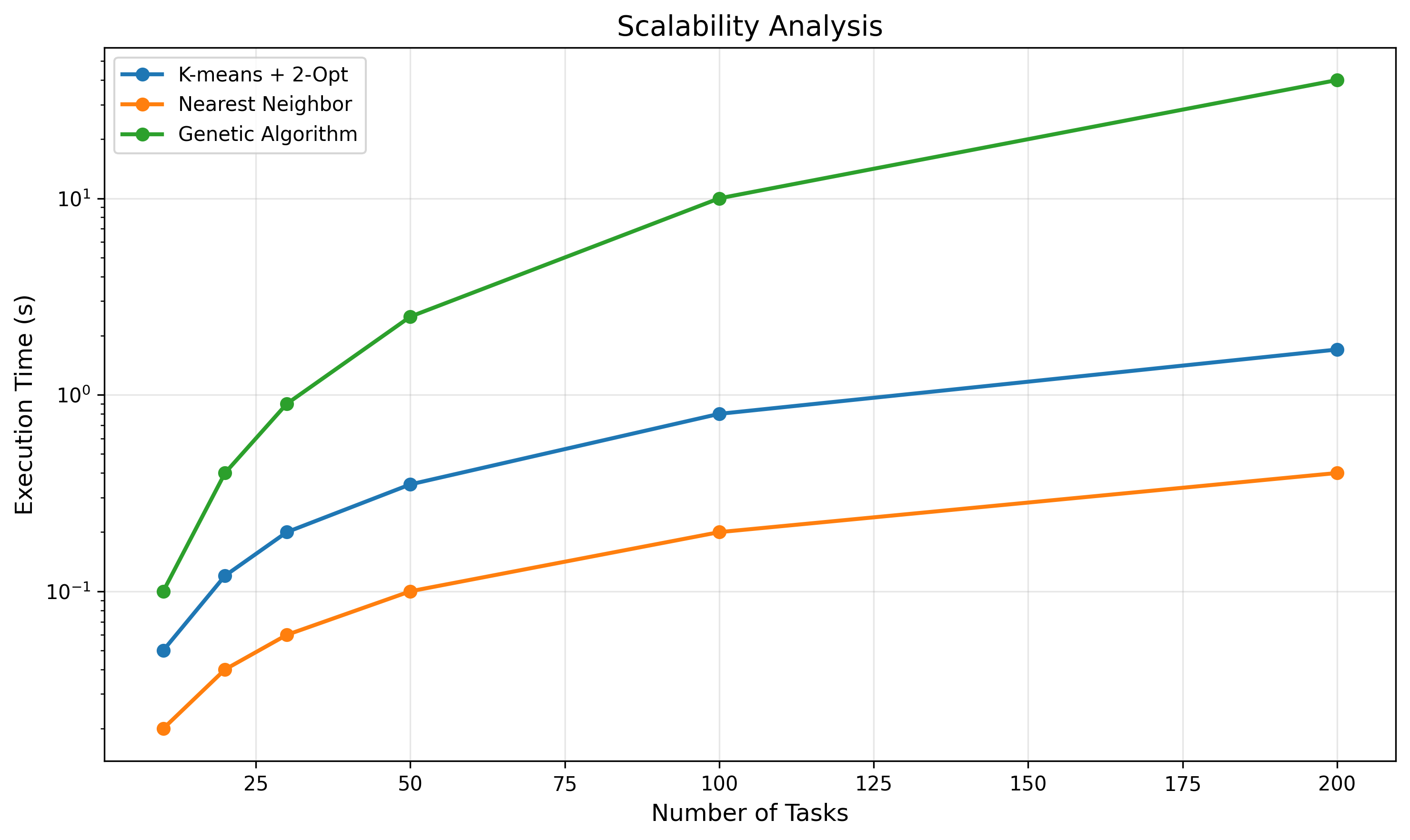}
\caption{Scalability Analysis}
\label{fig:scalability}
\end{figure}

We experimentally assess scalability by repeating tasks from 20 to 200 while keeping four robots, and evaluating with varying team size (3 and 5 robots). Figure \ref{fig:scalability} shows the computational time scaled across different method for different problem sizes. We have compared our K-means based 2-Opt Localization approach with global optimization methods. It is evident from the figure that the proposed model scale better with the total problem size than the GA based approach \cite{shao2021robot}. Our suggested 2-Opt algorithm also benefits largely when applied to smaller clusters, thus suitable for very well large scale applications consisting of many similar tasks.

\subsection{Limitations}\label{sec46}
Even though we have shown the effectiveness of our spatial clustering approach in performing task allocation and collision avoidance, there are limitations that should be considered. 
\begin{itemize}
    \item Clustering quality plays a heavy role in performance as K-means is sensitive to initial selection of centroids. 
    \item In dynamic scenarios, our assumption of a static environment poses difficulty for meeting the needs of dynamically emerging tasks as they might entail out-of-pocket re-clustering that could interfere with ongoing operations.
    \item Our approach may not be empirically effective for collision avoidance in constrained environments. Additional low-level collision avoidance mechanisms might be required in order for robots to traverse overlapping regions near the depot.
\end{itemize}


\section{Conclusion and Future Work}\label{sec5}

In this paper, we propose a methodology to solve the task allocation problem in the multi robot inspection scenarios by use of spatial clustering that jointly solves the task allocation and collision avoidance problems. The proposed approach based on K-means clustering for spatial partitioning and 2-Opt algorithm for route optimization inside clusters achieves rather high computational efficiency without any chance of collisions. Our strategy demonstrates that computational time can be reduced up to 93\% relative to other approaches, its solution quality is within 7\% of best performing method, and there are no inter robot intersections.

This work opens several promising research directions including theoretical analysis of collision avoidance properties and size of optimality gaps; design of dynamic and distributed implementations to handle variations in task arrival and robots failures; extensions to heterogeneous task scenarios with different level of sensor requirements; combining machine learning techniques for adaptive clustering; and comprehensive validation on physical robot platforms in realistic environments. At the same time, they would mitigate existing dependencies of presence of clustering quality on task distribution, adaptation capabilities to environment changes, guarantees of avoiding collisions over overlapping regions, and intrinsic optimality gap, that follow from problem decomposition. These advancements will allow for substantial improvement in the task allocation capability and applicability for the clustering based multi robot task allocation for a wide range of domains which require coordinated operations.


\end{document}